\documentclass[conference, twoside]{IEEEtran}
\IEEEoverridecommandlockouts
\usepackage{cite}
\usepackage{amsmath,amssymb,amsfonts}
\usepackage{algorithmic}
\usepackage{graphicx}
\usepackage{textcomp}
\usepackage{xcolor}
\def\BibTeX{{\rm B\kern-.05em{\sc i\kern-.025em b}\kern-.08em
    T\kern-.1667em\lower.7ex\hbox{E}\kern-.125emX}}

\usepackage{xspace}

\usepackage{lipsum}
\usepackage[nolist]{acronym}
\begin{acronym}

\acro{cnn}[CNN]{Convolutional Neural Network}
\acro{mllm}[MLLM]{Multi-Modal Large Language Model}
\acro{vlm}[VLM]{Vision-Language Model}
\acro{lvlm}[LVLM]{Large Vision-Language Model}

\acro{jde}[JDE]{Joint Detection and Embedding}
\acro{sde}[SDE]{Separate Detection and Embedding}
\acro{iou}[IoU]{Intersection over Union}
\acro{tbd}[TBD]{Tracking-by-Detection}
\acro{reid}[ReID]{Re-Identification}
\acro{mot}[MOT]{Multi-Object Tracking}
\acro{roi}[RoI]{Region of Interest}
\acro{giou}[GIoU]{Generalized Intersection over Union}
\acro{fifo}[FIFO]{First-In, First-Out}
\acro{mgn}[MGN]{Multi-Granularity Network}
\acro{bn}[BN]{Batch Normalization}
\acro{hota}[HOTA]{Higher Order Tracking Accuracy}
\acro{mota}[MOTA]{Multiple Object Tracking Accuracy}
\acro{vit}[ViT]{Vision Transformer}
\acro{dl}[DL]{Deep Learning}
\acro{ai}[AI]{Artificial Intelligence}
\acro{amr}[AMR]{Autonomous Mobile Robot}
\acro{map}[mAP]{mean Average Precision}
\acro{fps}[FPS]{Frames Per Second}
\acro{ema}[EMA]{Exponential Moving Average}

\end{acronym}
\usepackage{color,soul}
\usepackage{svg}
\usepackage{booktabs}
\usepackage{balance}
\usepackage{soul}
\usepackage{censor}
\usepackage{hyperref}
\usepackage{fancyhdr}

\begin{document}
% --- Headers ---
\pagestyle{fancy}
\fancyhf{}

\fancyhead[LE,RO]{\footnotesize \thepage}
\fancyhead[LO,RE]{\footnotesize PREPRINT VERSION.}

\renewcommand{\headrulewidth}{0pt}
% --- Headers ---

\title{Zero-Shot Semantic Re-Identification for Autonomous Driving: A VLM Baseline Study}

\author{
\IEEEauthorblockN{
Eduardo~Borges,
Manuel~Abreu,
Luís~Garrote,
Urbano~J.~Nunes
}
\IEEEauthorblockA{
Institute of Systems and Robotics, University of Coimbra, Coimbra, Portugal\\
\small\texttt{\{eduardo.borges, manuel.abreu, garrote, urbano\}@isr.uc.pt}
}
}

\maketitle
\thispagestyle{fancy}

\begin{abstract}
Re-Identification (ReID) in autonomous driving is typically formulated as a visual matching problem, where observations of vehicles, pedestrians, and cyclists are associated across time, frames, or camera views using learned appearance embeddings, often complemented by motion, geometric, or multimodal cues. However, purely visual representations may be sensitive to viewpoint, occlusion, illumination, and sensor-domain variations, limiting their interpretability and robustness in complex driving scenes. 
We propose a baseline study of a zero-shot pipeline using Vision-Language Models (VLMs) to generate textual descriptions of detected traffic participants and evaluate whether these descriptions can support identity matching across observations. Instead of relying only on low-level visual similarity, the proposed formulation represents each object through structured semantic attributes, including category, color, shape, pose, visible parts, spatial context, and distinctive visual cues.
This study provides an initial benchmark for language-based re-identification in autonomous-driving scenarios, discussing and evaluating the strengths and limitations of current VLMs for this task. 
Results demonstrate that zero-shot semantic descriptions can support effective object re-identification, achieving retrieval performance comparable to a supervised CNN baseline while offering greater interpretability through explicit identity cues. However, the experiments also reveal important challenges, including attribute inconsistency across viewpoints and limited fine-grained discrimination between visually similar instances.
\end{abstract}

\begin{IEEEkeywords}
Object Re-Identification, Vision-Language Models, Autonomous Driving, Semantic Representation, Zero-Shot.
\end{IEEEkeywords}

\section{Introduction}
Reliable perception is a fundamental requirement for autonomous driving systems. Beyond detecting traffic participants in individual frames, autonomous vehicles must maintain a consistent understanding of surrounding agents over time. This includes associating vehicles, pedestrians, cyclists, and other road users across consecutive observations, across multiple cameras, or across temporally separated viewpoints. This association problem, commonly referred to as object re-identification, plays an important role in tracking, behavior prediction, scene understanding, and long-term situational awareness.

Most re-identification methods in autonomous-driving scenarios formulate the problem as visual matching. Given two or more object crops, the objective is typically to learn an appearance-based embedding in which observations of the same object are close to each other, while observations of different objects are separated. This paradigm has led to significant progress, particularly with deep metric learning and transformer-based visual representations. However, autonomous-driving environments present several challenges that make purely visual re-identification difficult. Objects may be observed from different viewpoints, partially occluded by other traffic participants, affected by motion blur, or captured under changing illumination and weather conditions. In addition, the same object may appear differently across sensors, cameras, or time, while different objects may share very similar visual appearances.

\begin{figure}
    \centering
    \includegraphics[width=1\linewidth]{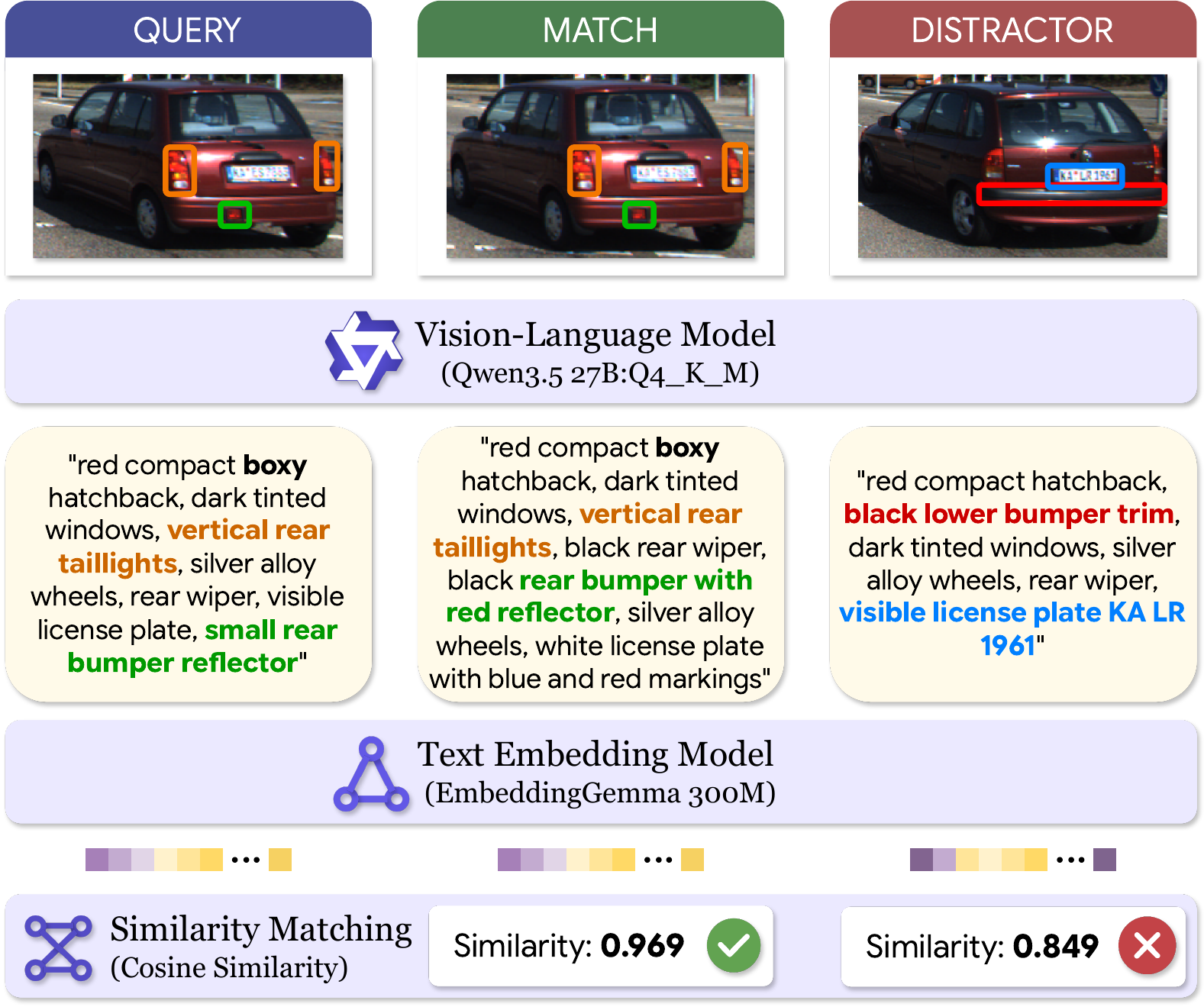}
    \caption{\textbf{Overview of the proposed zero-shot semantic re-identification pipeline.} Our pipeline consists of three stages: (1) generating discriminative one-line descriptions from object image crops using a VLM, (2) encoding the generated descriptions into text embeddings, and (3) retrieving objects via cosine similarity between embeddings.}
    \label{fig:paper_teaser}
\end{figure}

Recent advances in \acp{vlm} have opened new possibilities for connecting visual perception with natural-language descriptions. These models can generate textual descriptions of objects and scenes, answer questions about visual content, and extract semantic attributes from images. In the context of autonomous driving, VLMs offer the possibility of representing traffic participants not only as image embeddings, but also as structured semantic descriptions. Such descriptions may include object category, color, shape, pose, visible components, contextual information, and distinctive visual attributes. This raises an important question: can language-based semantic identity descriptions support re-identification in autonomous-driving scenarios? 
A baseline study on semantic identity descriptions for autonomous-driving re-identification is proposed in this work to assess the feasibility of using VLM-generated descriptions as an interpretable representation for identity matching.  These descriptions are compared across observations to assess whether they preserve sufficient semantic information to support re-identification.
Also, rather than replacing visual embeddings, semantic identity descriptions can serve as an additional representation layer that can support more transparent, human-interpretable, and potentially more robust perception systems for autonomous driving.

The motivation for this study is twofold. First, semantic descriptions may improve interpretability by making association identity cues more transparent. This can help diagnose failures, understand ambiguous cases, and support human-in-the-loop analysis of autonomous-driving perception systems. Second, language-based representations may complement conventional visual pipelines by capturing attributes that remain stable across moderate viewpoint or illumination changes.

The main contribution of this work is a zero-shot baseline study on autonomous-driving \ac{reid} that uses \ac{vlm}-generated descriptions as an interpretable intermediate representation, which should serve as support for research on multimodal and explainable re-identification, as well as applications in MOT. 

\section{Related Work}

\subsection{Object Re-Identification in Autonomous Driving}
In autonomous driving, re-identification is typically formulated as a problem of matching visual observations across time, with most approaches learning appearance embeddings directly from images through supervised metric learning. Re-Identification is typically integrated into \ac{mot} systems, which are responsible for detecting objects and maintaining consistent identities over time and commonly rely on object motion and/or appearance embeddings to maintain identity across frames.

DeepSORT~\cite{deepsort} extends SORT~\cite{sort} by integrating a deep association metric that uses nearest-neighbor matching in appearance space to improve long-term association and lower identity switches under occlusion. More recent \ac{mot} trackers such as BoT-SORT~\cite{botsort} combine motion and appearance cues with camera-motion compensation and an improved Kalman filter state vector, while StrongSORT~\cite{strongsort} revisits the DeepSORT baseline and strengthens detection, feature embedding, and trajectory association with lightweight plug-and-play modules.  

Transformer-based \ac{reid} models further improved object-level identity matching by learning richer global and part-level features. TransReID~\cite{transreid} replaces conventional \ac{cnn} backbones with a transformer-based architecture and reports state-of-the-art performance on both person and vehicle \ac{reid} benchmarks, showing that transformer representations are highly effective for fine-grained identity recognition.

These methods show that appearance cues are beneficial for tracking, but are dependent on learned visual representations.

\subsection{Language-Guided Re-Identification}
Language-based retrieval introduced a different formulation of identity matching, where a natural-language description serves as the query. This direction was established by the CUHK-PEDES dataset~\cite{CUHK-PEDES} and its associated person-search setting, by collecting large-scale pedestrian images paired with detailed textual descriptions. Early work in this area proposed attention-based retrieval models that rank gallery images based on the query text, demonstrating that semantic descriptions can encode useful identity cues.

More recently, some works have explored how pre-trained foundation models, such as CLIP~\cite{CLIP}, can be adapted to \ac{reid} without requiring explicit textual annotations for every identity. CLIP-ReID~\cite{clipreid} shows that CLIP can be leveraged for image \ac{reid} even without text labels by learning ID-specific prompt tokens in a first stage and then using those learned textual tokens to constrain image-encoder fine-tuning in a second stage. In PromptSG~\cite{PromptSG}, the authors argue that the usage of pre-defined prompts by CLIP-ReID constitutes an important limitation. As such, they propose overcoming this limitation by using a textual inversion technique to learn a pseudo-text token that aligns with the query image context.

This line of work suggests that while language priors can improve \ac{reid}, most current methods still rely on task-specific training or prompt tuning rather than a fully zero-shot pipeline.

\subsection{Vision-Language Models for Re-Identification}
Recent work has begun to exploit \acp{lvlm} more directly for \ac{reid}. LVLM-ReID~\cite{LVLM-REID} introduces a semantic token generation strategy in which an  \ac{lvlm} is instructed to produce a pedestrian semantic token that captures key appearance semantics from an image. This token is then refined through a semantic-guided interaction module that exchanges information between semantic and visual tokens, and the resulting representation is used for identity matching.

The authors of ChatReID~\cite{chatreid} push this idea further by treating person \ac{reid} as an open-ended interactive retrieval problem. It introduces a hierarchical progressive tuning strategy with three stages: person attribute understanding, fine-grained image retrieval, and multi-modal reasoning. In this framework, the model is progressively tuned from attribute-level semantics to retrieval-level discrimination, enabling visual question answering style interaction and strong performance across multiple \ac{reid} benchmarks.

Beyond standard single-modality matching, \acp{mllm} have also been leveraged to generate descriptive captions that explicitly guide multi-modal \ac{reid} architectures (e.g., fusing RGB, Near-Infrared, and Thermal images). For instance, NEXT~\cite{NEXT} employs MLLMs to generate attribute-aware captions, which are then used to text-modulate semantic experts that capture fine-grained, inter-modality complementary cues. Similarly, IDEA~\cite{IDEA} develops a standardized MLLM-based pipeline to extract structured attribute descriptions for multi-modal images. To mitigate background noise and fusion conflicts, IDEA introduces an Inverted Multi-modal Feature Extractor that inverts the generated global text into pseudo-image tokens, providing rich semantic guidance for subsequent deformable aggregation.

Overall, these methods suggest that semantic information from language and \acp{vlm} can improve identity matching, but they typically depend on supervised tuning or learned prompt components. In contrast, our work investigates a zero-shot, prompt-driven semantic signature pipeline for autonomous-driving object \ac{reid}, where a \ac{vlm} generates a compact identity-oriented description that is then embedded and compared across frames.

\section{Methodology}
\subsection{Overview}
Our pipeline consists of three stages, as presented in Fig.~\ref{fig:paper_teaser}. First, each labeled object instance in KITTI is cropped from the source image to form a single-object input. Second, a \ac{vlm} generates a one-line description of the crop using a prompt designed for identity discrimination. Third, the generated text is encoded with a sentence embedding model, and retrieval is performed by cosine similarity between object embeddings.

Let $x_i$ denote the cropped image for object instance \(i\). A \ac{vlm} $f_{vlm}$ produces a text description
\begin{equation}
    t_i = f_{vlm}(x_i,p)
\end{equation}
where $p$ is the retrieval-oriented prompt. A text encoder $f_{text}$ then maps the description to an embedding
\begin{equation}
    \mathbf{e}_i = f_{text}(t_i) \in \mathbb{R}^d
\end{equation}
At evaluation time, a query object is compared with a gallery of objects using cosine similarity between embeddings:
\begin{equation}
s(q,g)=
\frac{\mathbf{e}_q^\top \mathbf{e}_g}
{\|\mathbf{e}_q\|_2 \, \|\mathbf{e}_g\|_2} \in [-1,1]
\end{equation}
where \(\mathbf{e}_q\) and \(\mathbf{e}_g\) denote the query and gallery embeddings, respectively. Gallery objects are ranked according to their similarity score \(s(q,g)\), with higher values indicating greater semantic similarity.

\subsection{Dataset}

Experiments are conducted on KITTI-ReID, a benchmark previously introduced in our earlier work~\cite{borges2026appearance}. KITTI-ReID is derived from the KITTI tracking dataset by extracting object crops from the annotated bounding boxes and associating them with their corresponding identity labels. Each sample, therefore, consists of a single-object image crop together with its class and identity annotation, enabling instance-level retrieval and re-identification evaluation. Additional details regarding dataset construction, statistics, and annotation processing are provided in~\cite{borges2026appearance}.

The dataset preserves the original temporal identity associations provided by KITTI while removing most scene context, forcing retrieval methods to rely primarily on object appearance. This design makes KITTI-ReID particularly suitable for studying object-centric representations and assessing the discriminative power of alternative embedding strategies, including the language-based representations explored in this work.

\subsection{Prompting Strategy}
The prompt is designed to elicit compact descriptions with high instance-level discriminability. It explicitly instructs the \ac{vlm} to:

\begin{enumerate}
    \item Produce exactly one line of plain text,
    \item Avoid hedging language,
    \item Omit uncertain or invisible details,
    \item Prioritize stable identity-bearing attributes,
    \item Follow a fixed attribute order,
    \item And emphasize concrete visual cues such as shape, color, trim, markings, accessories, and wear.
\end{enumerate}

The prompt is domain-aware. For vehicles, it prioritizes body type, silhouette, trim, windows, wheel design, lights, decals, and visible damage. For pedestrians, it prioritizes clothing, hair, eyewear, headwear, bags, footwear, and other visible accessories. This structured prompting aims to reduce variability in the textual representation while still allowing the model to describe salient details.

\subsection{Model Selection}
To evaluate the feasibility of language-mediated \ac{reid} in a strictly zero-shot setting, all models are used off-the-shelf without any task-specific fine-tuning, prompt tuning, adapter training, or parameter updates. 

We evaluate four \acp{vlm}:
\begin{itemize}
    \item Qwen3.5 0.8B~\cite{qwen3.5}
    \item Qwen3.5 9B (Q4-K-M)~\cite{qwen3.5}
    \item Qwen3.5 27B (Q4-K-M)~\cite{qwen3.5}
    \item Gemma 4 E2B~\cite{gemma4}
\end{itemize}

We pair these with four text embedding models:
\begin{itemize}
    \item Embedding Gemma 300M~\cite{embedgemma}
    \item Nomic Embed v2 305M~\cite{nomicembedv2}
    \item Qwen3-Embedding 4B~\cite{qwen3embedding}
    \item Qwen3-Embedding 8B~\cite{qwen3embedding}
\end{itemize}

This yields a cross-product of description generators and encoders, allowing us to isolate the impact of semantic description quality from that of the embedding space. The resulting evaluation thus reflects the capabilities of current foundation models in a purely inference-based, zero-shot \ac{reid} setting.

\begin{figure*}[t]
    \centering
    \includegraphics[width=1\textwidth]{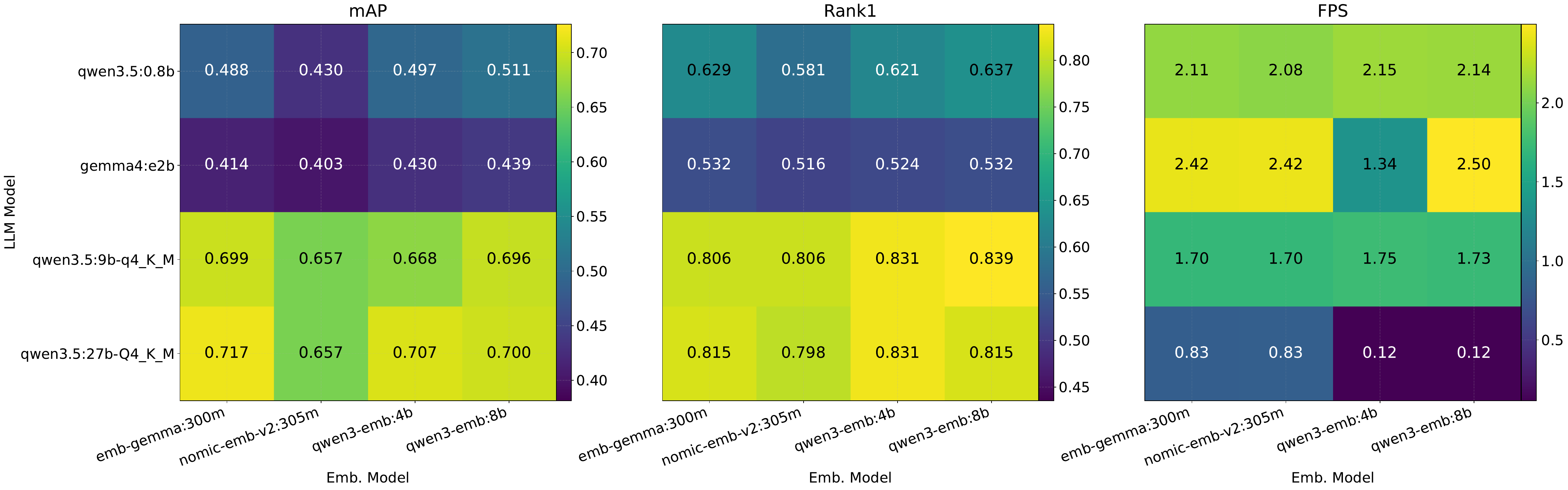}
    \vspace{-2em}
    \caption{\textbf{Multi-metric evaluation (mAP, Rank-1, FPS) of the proposed pipelines.} The data isolates the impact of specific core LLMs and embedding models on both system accuracy and whole-system inference latency.}
    \label{fig:heatmaps}
    \vspace{-1em}
\end{figure*}

\subsection{Retrieval and Evaluation}

After embedding all object descriptions, retrieval is performed by ranking gallery items according to cosine similarity with the query embedding. We follow the standard re-identification evaluation protocol used in the Market-1501~\cite{Liang2015market1501} benchmark and report:

\begin{itemize}
    \item Mean Average Precision (mAP), measuring the quality of the full ranked list.
    \item Rank-1 accuracy, measuring whether the top retrieved match belongs to the correct identity,
\end{itemize}

In addition to retrieval accuracy, we evaluate inference efficiency. For each \ac{vlm}, we measure the average description generation throughput in frames per second (FPS) and the average processing latency per object crop. Since the proposed pipeline relies on language generation as an intermediate representation, computational efficiency is an important consideration for deployment in real-world autonomous driving systems. Reporting both accuracy and throughput enables analysis of the trade-off between description quality and runtime performance across models of different sizes. Reported runtime includes both description generation and embedding extraction, representing the end-to-end cost of the proposed language-mediated re-identification pipeline.

To ensure a fair comparison, all \acp{vlm} are evaluated using the same prompt, all text embeddings are generated without task-specific fine-tuning, and all measurements are performed on identical hardware. This setting isolates the effect of the language model and embedding model choices while assessing the practical viability of language-mediated object re-identification.

\section{Experiments}

\subsection{Experimental Setup}
All experiments were conducted on a workstation equipped with an AMD Ryzen 9 7900X CPU, 64GB of RAM, and a single NVIDIA RTX 5090 GPU. Vision-language inference was performed locally using Ollama.

We evaluate all combinations of the four VLMs and four text embedding models on KITTI-ReID. For each crop, the \ac{vlm} generates a description according to the fixed prompt, and the resulting text is encoded into a vector representation. Retrieval is then computed using cosine similarity. The main experimental variables are 1) the choice of \ac{vlm}, and 2) the choice of sentence embedding model.

Because the objective is to assess representation quality rather than to optimize a learned retriever, we keep the pipeline fully inference-based. This makes the results directly interpretable: improvements can be attributed to better textual descriptions, better text embeddings, or both.

\subsection{Retrieval Accuracy}
We evaluate the retrieval performance of the proposed text-mediated pipeline across all \ac{vlm} and embedding model combinations. The results in the accuracy heatmaps (Fig.~\ref{fig:heatmaps}) suggest the description-generating \ac{vlm}'s parameter scale is a major factor for instance-level discriminability.

The Qwen3.5-27B model achieves peak performance across all configurations, yielding a maximum mAP of 0.717 when paired with the EmbeddingGemma 300M encoder, and Rank-1 accuracy of 0.831 when paired with the Qwen3-Embedding-4B. The heavily quantized Qwen3.5-9B model demonstrates comparable robustness, incurring a marginal performance loss (0.699 mAP, 0.831 Rank-1) while maintaining the same embedding space.

Conversely, the smaller sub-3B parameter models exhibit a poorer performance in ReID capabilities. The Qwen3.5-0.8B achieves a maximum of 0.511 mAP, and the Gemma4-2B model falls to 0.439 mAP, when paired with the Qwen3-Embedding-8B. These results indicate that description quality is a limiting factor in the current pipeline, since improvements from the text encoder alone do not recover the performance lost by weaker \ac{vlm} outputs.
% This indicates that low-capacity \acp{vlm} might fail to capture or articulate the fine-grained, identity-bearing visual features required to distinguish inter-class objects in a pure textual domain.

\subsection{Impact of the Text Embedding Space}
While the \ac{vlm} defines the semantic upper bound of the pipeline, the sentence embedding model determines how effectively these semantic descriptions are organized within the retrieval metric space. The results show that increasing the scale of the embedding model does not necessarily translate into higher retrieval accuracy, suggesting that model size alone is not sufficient to improve identity matching performance.
In fact, the highest overall performance is achieved using the smallest tested encoder: the 300M-parameter Embedding-Gemma model yields a peak mAP of 0.717  when paired with the Qwen3.5-27B \ac{vlm}. In contrast, applying the significantly larger 8B-parameter Qwen3-Embedding model to the exact same 27B \ac{vlm} outputs reduces the mAP to 0.700. This indicates that the alignment between the \ac{vlm}'s specific generated vocabulary and the encoder's inherent training distribution is a stronger predictor of \ac{reid} performance than the raw parameter scale of the embedding model.

However, the results suggest that description quality remains the dominant limiting factor: an embedding model, regardless of its alignment or capacity, cannot synthesize visual details omitted during the generation phase. Across all tested configurations, improvements in the embedding model did not fully compensate for the weaker descriptions produced by the lower-capacity \acp{vlm}.

\subsection{Efficiency and Runtime Trade-offs}
\begin{figure}[tb]
    \centering
    \includegraphics[width=1\linewidth]{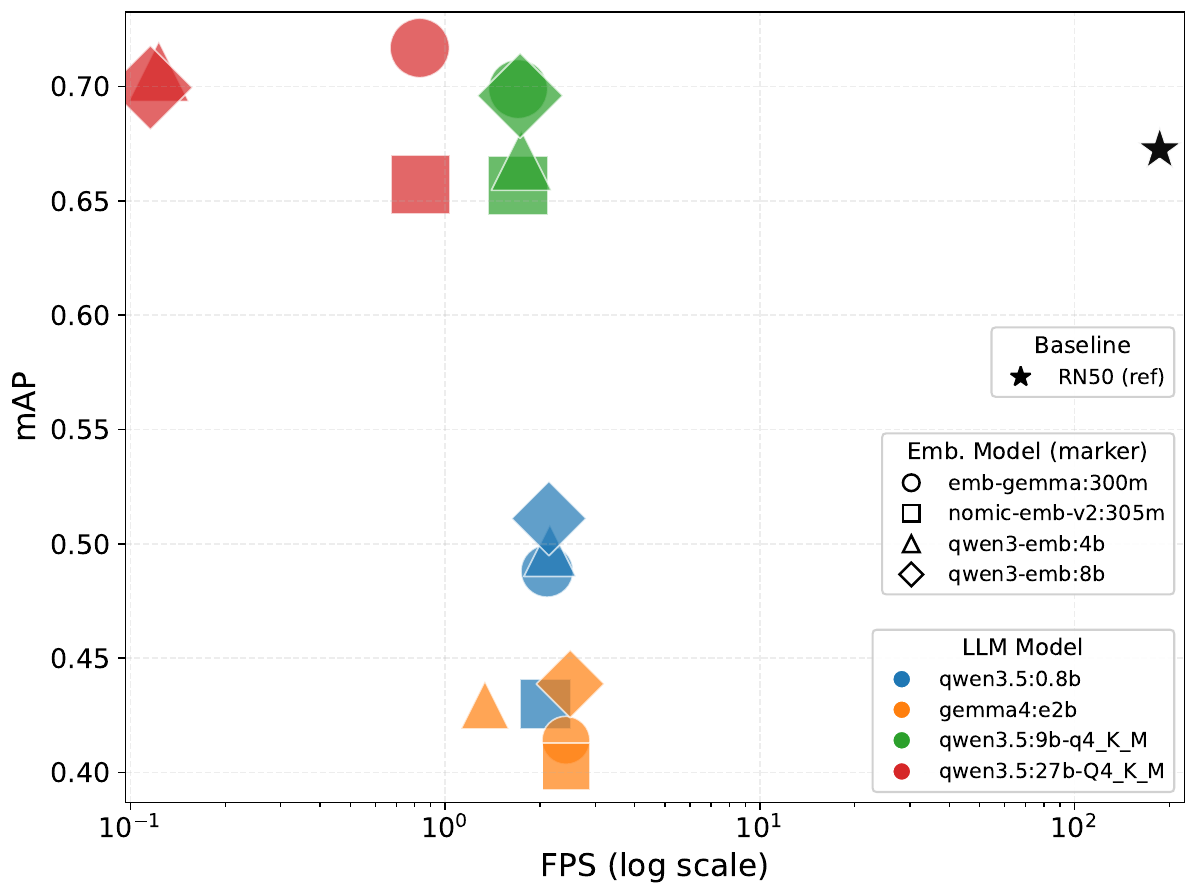}
    \vspace{-2em}
    \caption{\textbf{Accuracy--speed trade-off on KITTI-ReID.} Marker size is proportional to Rank-1 accuracy. The figure compares zero-shot VLM-based pipelines and a supervised ResNet50 baseline pretrained on Market-1501, illustrating the trade-off between retrieval performance (mAP) and inference throughput (FPS).}
    \label{fig:trade_off}
\end{figure}
For autonomous driving applications, inference latency dictates the practical utility of perception pipelines. Fig.~\ref{fig:trade_off} illustrates the trade-off between retrieval accuracy (mAP and Rank-1) and throughput (FPS). Runtime measurements represent the end-to-end latency \textit{per object crop}, dominated primarily by the autoregressive text generation of the \ac{vlm}. As a supervised reference point, we ran a ResNet50~\cite{ResNet} baseline pretrained on Market-1501 which achieved 0.6724 mAP, 0.8548 Rank-1, and 186.68 FPS, placing it at the high-efficiency end of the spectrum and providing a strong non-LLM comparison for the proposed method.

The accuracy–speed trade-off highlights the operational limitations of the largest models. Despite achieving peak accuracy, the 27B quantized model operates at approximately 0.12 to 0.83 FPS, suggesting that it is better suited to offline auto-labeling or data mining. The 9B quantized model has increased viability, retaining high discriminability while operating at approximately 1.70 to 1.75 FPS. The smallest models (0.8B and 2B) achieve the highest throughput among the VLM-based methods (2.08 to 2.43 FPS) but suffer near 30\% drops in mAP, failing the accuracy requirements for reliable identity association.

The horizontal clustering of different shapes for any given \ac{vlm} suggests that the choice of text encoder contributes little additional latency compared with the \ac{vlm} stage. The computational bottleneck resides primarily in the visual-language decoding phase. In contrast, the ResNet50 baseline demonstrates the efficiency of a conventional CNN-based ReID approach, achieving substantially higher throughput at the cost of reduced semantic flexibility. This suggests that future improvements for real-time text-mediated \ac{reid} must either accelerate \ac{vlm} inference or use this explicit text generation pipeline offline to distill descriptive capabilities into faster, single-pass feedforward architectures.

\subsection{Qualitative Analysis}
\begin{figure*}
    \centering
    \includegraphics[width=1\linewidth]{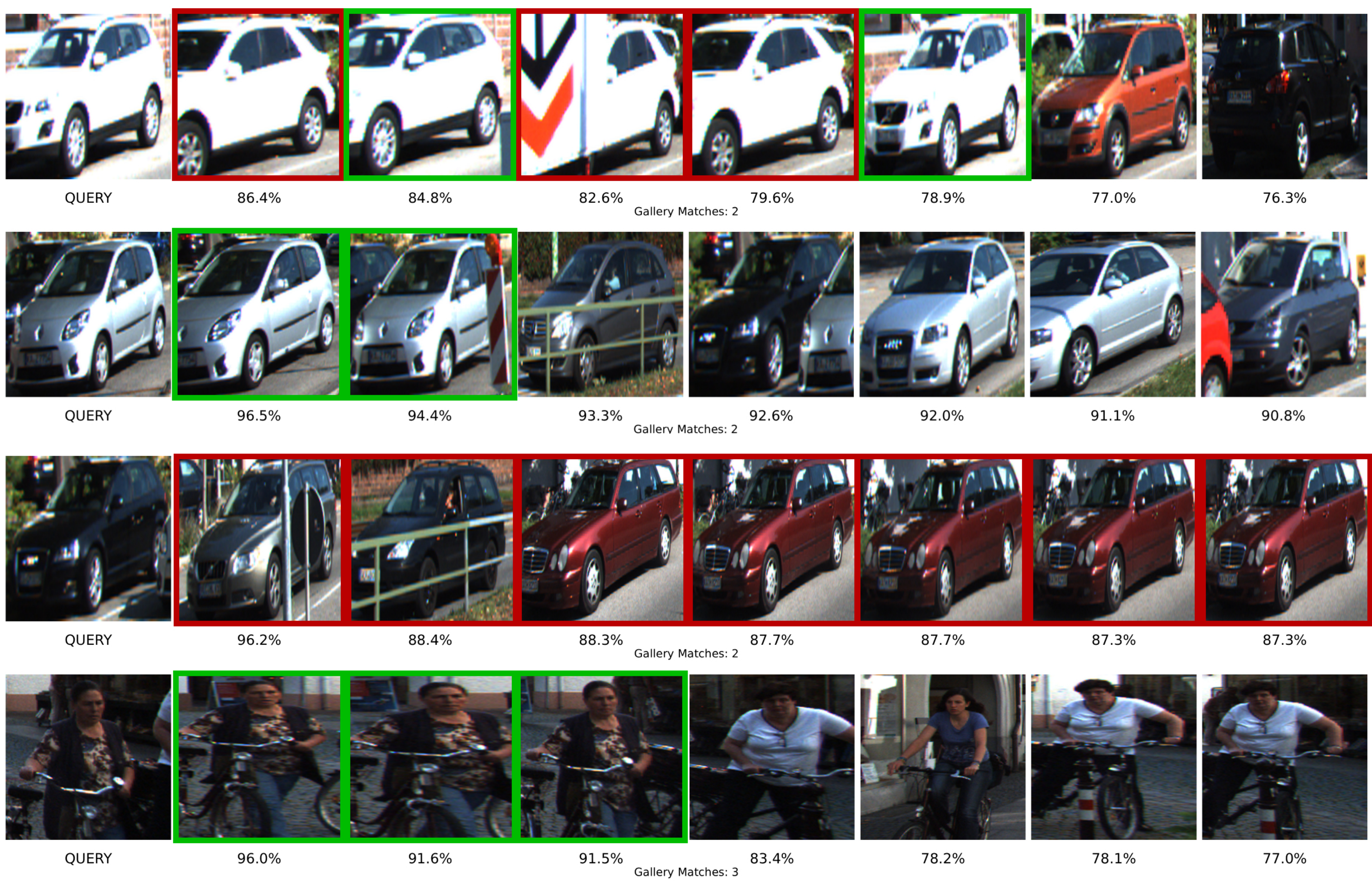}
    \vspace{-2em}
    \caption{\textbf{Qualitative retrieval examples on KITTI-ReID using the proposed zero-shot semantic re-identification pipeline.} For each query (left), 7 top-ranked gallery results are shown with cosine-similarity scores. Green boxes indicate correct identity matches, while red boxes denote distractors. The examples illustrate that the generated visual signatures can capture discriminative cues needed for matching across viewpoint changes and different object classes.}
    \label{fig:qualitative}
    \vspace{-1em}
\end{figure*}

Fig.~\ref{fig:qualitative} shows representative retrieval results on KITTI-ReID for both vehicles and pedestrians. In many cases, the proposed zero-shot semantic representation retrieves the correct identity among the highest-ranked gallery entries. The qualitative examples highlight that the generated descriptions preserve fine-grained, identity-bearing cues such as vehicle body shape, wheel design, trim, scratches, and plate-related details, as well as pedestrian clothing, hair style, and carried items. These cues are particularly useful when multiple objects share similar coarse appearance (such as chassis style, clothing, and accessories), since they help separate visually close instances that would otherwise be difficult to distinguish using only category-level information. At the same time, the figure also reveals that performance can degrade when the object is heavily occluded or when only limited visual detail is visible, which motivates future work on multimodal fusion and integration into a full MOT pipeline.

\section{Conclusion}
We introduced a zero-shot, text-mediated re-identification baseline for autonomous driving, where a \ac{vlm} produces semantic descriptions of object crops and a text encoder performs identity matching in language space. The results demonstrate that, even without training, the approach can achieve strong retrieval performance on KITTI-ReID and that semantic descriptions can preserve useful identity information for re-identification. However, the experiments also show a clear accuracy-efficiency trade-off: the best-performing models are too slow for strict real-time deployment in our setting, while the smaller models are faster but significantly less discriminative. As a result, the current pipeline is most suitable for offline processing, auto-labeling, and parallel verification alongside a conventional \ac{reid} system. Future work will focus on accelerating the generation stage, and integrating semantic descriptions with standard visual re-identification and MOT pipelines.

\section*{Acknowledgment}
This work has been supported by the Portuguese Foundation for Science and Technology (FCT) through grant ISR-UC UID/00048/2025 (DOI: 10.54499/UID/00048/2025) and by Agenda ``GreenAuto: Green innovation for the Automotive Industry", with reference 02/C05-i01.01/2022.PC644867037-00000013. Eduardo Borges is being supported by the FCT Ph.D. grant 2025.01736.BD.

\balance
\bibliographystyle{IEEEtran}
\bibliography{refs.bib}

\end{document}